\documentclass[10pt,twocolumn]{article}

\usepackage{wacv}
\usepackage{times}
\usepackage{epsfig}
\usepackage{graphicx}
\usepackage{amsmath}
\usepackage{amssymb}
\usepackage{floatrow}

\usepackage[pagebackref=true,breaklinks=true,letterpaper=false,colorlinks,bookmarks=false]{hyperref}

\wacvfinalcopy 

\ifwacvfinal\fi

\newcommand{\comment}[1]  {  }

\newcommand{\cH}{\mathcal{H}}
\newcommand{\cP}{\mathcal{P}}

\newcommand{\bq}{{\mathbf q}}
\newcommand{\bp}{{\mathbf p}}
\newcommand{\bP}{{\mathbf{P}}}

\newcommand{\sH}{{\mathsf{H}}}
\newcommand{\sh}{{\mathsf{h}}}
\newcommand{\sD}{{\mathsf{D}}}
\newcommand{\sd}{{\mathsf{d}}}
\newcommand{\sll}{{\mathsf{l}}}
\newcommand{\slL}{{\mathsf{L}}}

\newcommand{\figs}[1]{figs/}

\begin{document}

\title{A Cheap System for Vehicle Speed Detection}

\author{Chaim Ginzburg, Amit Raphael and Daphna Weinshall\\
School of Computer Science and Engineering, Hebrew University of Jerusalem, Israel\\
{\tt\small daphna@cs.huji.ac.il}
}

\maketitle

\begin{abstract}

The reliable detection of speed of moving vehicles is considered key to traffic law enforcement in most countries, and is seen by many as an important tool to reduce the number of traffic accidents and fatalities. Many automatic systems and different methods are employed in different countries, but as a rule they tend to be expensive and/or labor intensive, often employing outdated technology due to the long development time. Here we describe a speed detection system that relies on simple everyday equipment - a laptop and a consumer web camera. Our method is based on tracking the license plates of cars, which gives the relative movement of the cars in the image. This image displacement is translated to actual motion by using the method of projection to a reference plane, where the reference plane is the road itself. However, since license plates do not touch the road, we must compensate for the entailed distortion in speed measurement. We show how to compute the compensation factor using knowledge of the license plate standard dimensions. Consequently our system computes the true speed of moving vehicles fast and accurately. We show promising results on videos obtained in a number of scenes and with different car models.

\end{abstract}

\section{Introduction}
\label{sec:intro}

\begin{figure}[htb]
\begin{center}
\includegraphics[width=.85\linewidth]{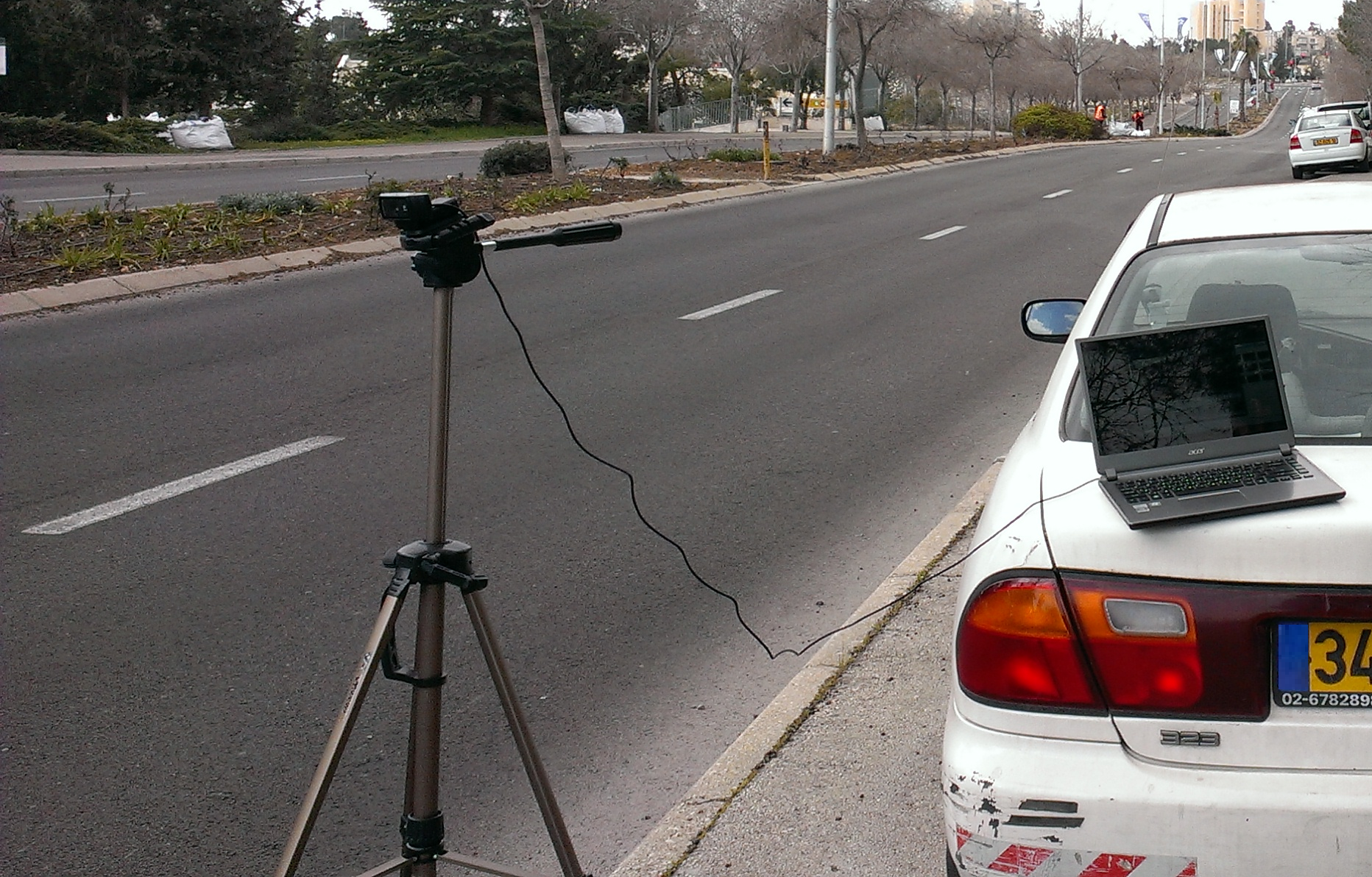} 
\end{center}
 \caption{A snapshot of our speed detection system, mounted in position to measure the speed of moving cars. }
 \label{fig:system}
\end{figure}

With the ever increasing number of cars worldwide, there is a growing need for cheaper and more
efficient automated traffic control systems. One important feature of such systems is the ability to
detect speed reliably - research seems to show that speed enforcement reduces the number of
accidents and the number of fatalities \cite{aarts2006driving,wilson2010speed}, thus saving
lives. Traditional speed detection devices require specific sensors like laser, radar, infrared or
ground sensors such as magnetic bars installed under the road, in addition to a camera which is
required in order to document offensive vehicles (see \eg \cite{Optotraffic,Brekford}).  Recently
computer vision technology has been used for the detection of speed based on stereo vision
\cite{KRIA} using multiple cameras.\footnote{We do not consider as comparable systems which compute
  the average speed traveled by a car between distant fixed sites.}  Our goal in this work is to
develop an automatic system that detects speed efficiently and reliably with cheap equipment, based
on a low-end laptop and a single consumer camera as illustrated in Fig.~\ref{fig:system}.

The most straightforward way to compute speed from a single RGB or monochrome stationary camera
would assume that the camera is fully calibrated, and therefore one can compute the $3D$ location of
every point in the image.  This system will track each car along the video frames in which the car
is visible. It will recover the exact location of the car when it had first appeared in the video,
and the last location before it had disappeared. It is now a simple matter to compute speed by
computing the distance the car had travelled, and dividing it by travel time which is determined by
the number of video frames between the end points.

The brute-force computation outlined above requires accurate camera calibration, including the
camera's exact location and orientation in $3D$ space, and its internal calibration parameters such
as zoom and focus (see review of calibration methods in \cite{zhang2001camera}). The accuracy of
camera calibration, however, is hard to guarantee for an autonomous camera \cite{zhang1997self},
even a stationary one, and partially for this reason such a system is not in common use.

Our method is based on the observation that full calibration is not necessary, given the constrained
environment under which the system needs to operate. In other words, since the system is required to
compute the speed of objects moving on a flat surface, one can use shortcuts and rely only on
partial plane calibration, which is easy to maintain and which is sufficient for the task. Such
partial calibration is sometimes called plane + parallax \cite{irani1996parallax}, or calibration to
a reference plane \cite{irani1998reference}. In Section~\ref{sec:speed_correction} we show that for
speed computation it is sufficient to calibrate the road only, which only guarantees the correct
recovery of the $3D$ location of points on the road.

Specifically, we assume that the road in the operational area of the system lies on an approximately
flat surface.  We call this surface the {\em reference plane}. In principle it is a simple matter to
compute a $2D$ projective transformation from the image plane to the {\em reference plane}, which
will map every point in the image which in the real world lies on the road to its real location in
$3D$ space \cite{faugeras1993three}. In order to compute this transformation, one needs to fix some
visible calibration pattern on the road. The planar calibration pattern should include at least 4
points. The larger it is, or the closer the calibration points are to the end of the visible
surface, the more robust the computation is (see example in Fig.~\ref{fig:calibration}).

Automatic camera calibration is often assisted by a calibration pattern presented to the camera.
However, when calibration is restricted to a $2D$ reference plane rather than the full $3D$ space,
the calibration pattern can be planar rather than 3-dimensional. In addition, the minimal number of
required calibration points is smaller. Thus calibration to a {\em reference plane} is more suitable
for the task of detecting speed of vehicles moving over a planar surface, and can be achieved more
readily.

Similar challenges were taken up in \cite{rad2010vehicle,liyanage2012homography}, for example, but
there are many implementation and other differences as compared to our work. For one, in
\cite{rad2010vehicle}, there is a need for a {\em test drive} with a known vehicle and speed in
order to evaluate the homography matrix. In \cite{liyanage2012homography}, in order to compute the
homography one needs some previously known distances on the road. This system requires some very
expensive equipment or the use of more than a single camera, because of the distance between the
camera position and the moving vehicles.  Finally, our system finds a specific spot on the vehicle
(the corner of the license plate), while the system in \cite{rad2010vehicle} only evaluates the {\em
  center of mass} of the moving object, and \cite{liyanage2012homography} uses a closing blob of the
object which is more error prone. Having said that, our system requires some knowledge about
standard dimensions of license plates, which is more readily available in isolated countries (such
as Iceland or Israel).

Next, we describe our method in Section~\ref{sec:method}. Experimental results are shown in
Section~\ref{sec:exps}.

\section{Method}
\label{sec:method}

Our task is to compute the speed of a car based on tracking its movement through a sequence of
images. Since in the end it is also necessary to identify the car, the natural choice of a target to
track is the car license plate. Our computation therefore starts by tracking the license plates of
moving cars, identifying the license number and its motion in image pixels. Subsequently it remains
to compute the real world motion of the car from its image motion.

Recall from the discussion in Section~\ref{sec:intro} that our method is based on calibration to a
{\em reference plane}.  Suppose the car has traveled from point $\bp_1$ to point $\bp_2$ in the
image.  Since the image itself is a plane, it is possible to compute a homography - a $3\times 3$
projective transformation of the $2D$ projective plane to itself - from the image plane to any other plane in
$3D$ space. In order to do this, one need correspondence between at least 4 marker points in the
image and their exact location on the real plane in $3D$. Four corresponding points define the
homography uniquely, while additional points can be used to make the computation more robust
\cite{hartley2003multiple}.

Suppose that we can track interest points on the car which touch the road, and that the road in the
surveyed area is planar. Now a natural choice for the real world target plane is the plane on which
the road lies (the {\em road plane}), and the road becomes {\em the reference plane}. By aligning
the image plane with {\em the reference plane} and tracking points on the image plane, the image
distance traveled by points on the road in the real world is equal to the actual distance traveled
by these points, with no need for any further camera calibration.

\comment{
\begin{SCfigure}[20][htb]
\includegraphics[width=0.6\linewidth]{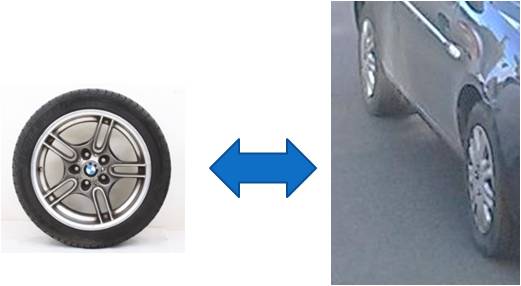}
\caption{Tracking the intersection of the wheel with the road can be difficult to do reliably.}
\label{fig:wheel}
\end{SCfigure}

\begin{figure}[htb]
\floatbox[{\capbeside\thisfloatsetup{capbesideposition={left,bottom},capbesidewidth=.27\linewidth}}]{figure}[\FBwidth]
{\hspace{0.1cm}\includegraphics[width=1.1\linewidth]{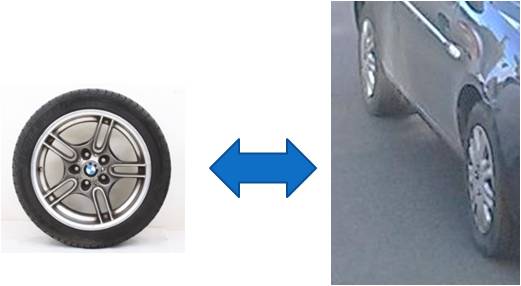}}
{\caption{Tracking the intersection of the wheel with the road can be difficult to do reliably.}
\label{fig:wheel}}
\end{figure}
}
\begin{figure}[htb]
\includegraphics[width=0.8\linewidth]{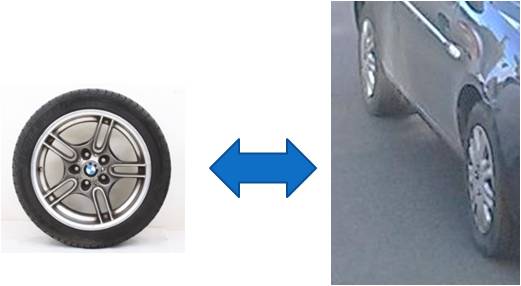}
\caption{Tracking the intersection of the wheel with the road can be difficult to do reliably.}
\label{fig:wheel}
\end{figure}

In practice, however, it is very difficult to track points on the car which touch the road, see
Fig.~\ref{fig:wheel}. The bottom part of wheels often lies in shadow, and a wheel's exact
intersection with the road is typically hard to locate consistently. On the other hand, it is rather
easy to track license plates, which have clear boundaries and often display a unique color. While
the license plate is also moving on a plane, it is not the road plane; rather it is a virtual plane
parallel to the road plane, whose distance from the road plane corresponds to the height of the
license plate over the road.

Thus, in order to compute the distance traveled in the real world from the distance in the image
plane projected to the road plane, we need to compute a correction factor that depends on the height
of the license plate over the road. If we know the car model, possibly by given access to a database
of all licensed cars via the license number, we can obtain this correction factor
directly. Otherwise, we describe in Section~\ref{sec:speed_correction} a method to compute this
correction factor from the distortion of the license plate itself, assuming that all license plates
adhere to some fixed standard size.

The final algorithm goes as follows: 
\begin{itemize}
\item {\em Pre-processing:} 
Compute the homography $\cH$ between the image plane and the road plane by identifying known markers
on the road which are visible to the camera. We rely on the fact that the exact location of the
markers on the road is measured and known apriori (Section~\ref{sec:homog}).
\item
Segment and track the license plate of each car in the image sequence for as long as possible
(Section~\ref{sec:OF}).
\item
Read the license number of the car from the license plate (Section~\ref{sec:NN}).
\item
Compute the actual distance traveled by the license plate in the real world, following these steps:
\begin{enumerate}
\item
Track the corner of the license plate through a sequence of frames, and identify its locations in
each frame. Project the locations onto the road plane using $\cH$\footnote{If necessary and if the
  markers are clearly visible, update the homography $\cH$ computed in the pre-processing
  step. Currently $\cH$ is updated every hour, although we have not observed any significant changes
  in $\cH$ in our experiments.} in order to obtain an estimate for the {\em real world} locations of
the corner. For each pair of frames in the sequence, calculate the difference between the locations
of the projected points and divide it by the time that passed between the frames, in order to obtain
several results for the estimation of the projected speed $s$ of the vehicle
(Section~\ref{sec:speed}).  To achieve robustness, return the median of the set of estimated speeds
as the final estimate for the motion of the license plate.
\item
Since the license plate is not located on the road in the real world, its projected estimated speed
$s$ is not its actual speed as explained above. In Section~\ref{sec:speed_correction} we describe
how to compute the correction factor $\rho$ which transforms $s$ to the license plane's (and
therefore the car's) actual speed $v=\rho s$.
\end{enumerate}
\item
Report the speed of the car as $v$. Identify the car by its license number.
\end{itemize}

\subsection{Computing the $2D$ calibration homography}
\label{sec:homog}

The projection of the $3D$ world to a $2D$ image via a pinhole camera can be elegantly expressed in
homogeneous coordinates as a linear transformation from $3D$ projective space to $2D$ projective
space. Specifically, let $\bP=[X,Y,Z,1]$ denote the homogeneous coordinates of a point in $3D$
projective space, and let $\bp=[x,y,1]$ denote the image homogeneous coordinates of the same point
viewed by a pinhole camera. Then there exists a $3\times 4$ matrix $\cP$ such that $\bp \propto \cP
\bP$. $\cP$ is the calibration matrix of the pinhole camera.

If all the points in space $\bP_i$ lie on some $3D$ plane, we can represent these points by their
relative $2D$ coordinates on the plane on which they lie $\bq_i$, where $\bq_i$ are vectors in the
$2D$ projective space. It now follows that there exists a homography $3\times 3$ matrix $\cH$ such
that
\begin{eqnarray}
\label{eq:homog}
\bp_i \propto \cH \bq_i ~\forall i
\end{eqnarray}
where $\propto$ denotes equality up to multiplication by a single scale factor. In (\ref{eq:homog})
there are 8 unknowns to recover (the elements of $\cH$ up to a scaling factor) and each point
provides 2 independent constraints on these unknowns. Therefore $\cH$ can be recovered from at least
4 corresponding points between the two planes, and specifically the correspondence $\{\bp_i\}$ to
$\{\bq_i\}$. In order for the computation to be robust, it is desirable that the points which are
used for obtaining $\cH$ lie as far as possible from each other towards the edges of the calibration
plane, see \cite{hartley2003multiple}.

\begin{figure}[htb]
\begin{center}
\includegraphics[width=.6\linewidth]{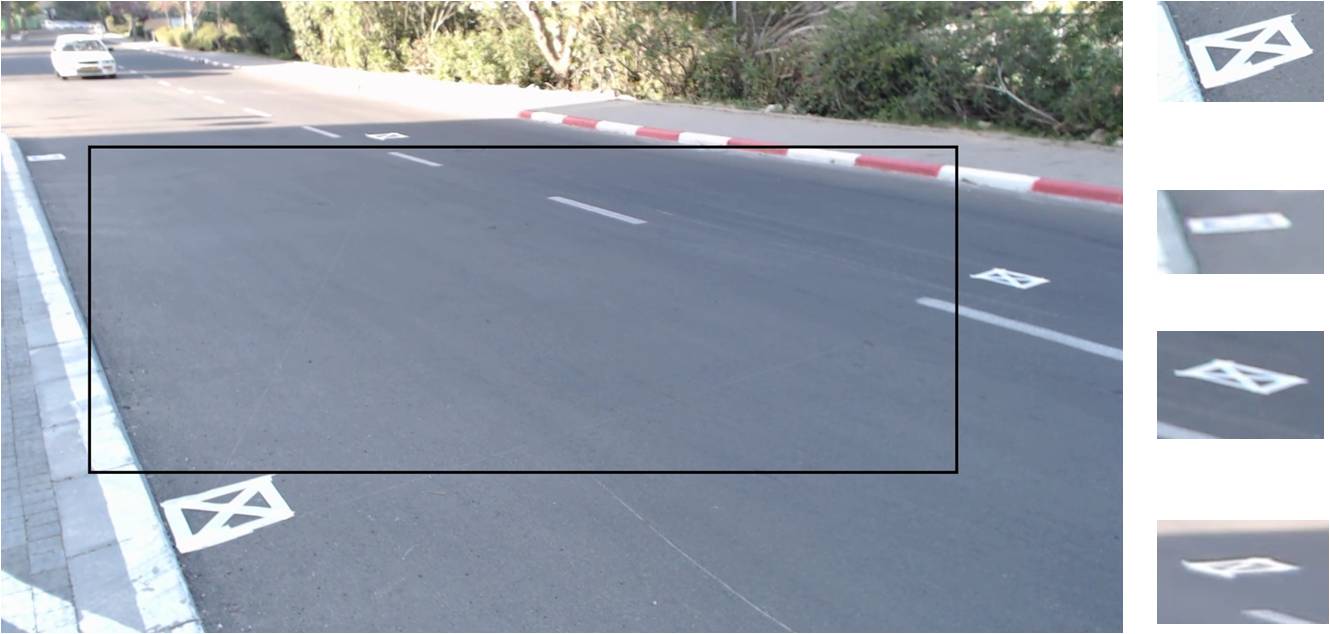} 
\hspace{0.75cm}
\includegraphics[width=.15\linewidth]{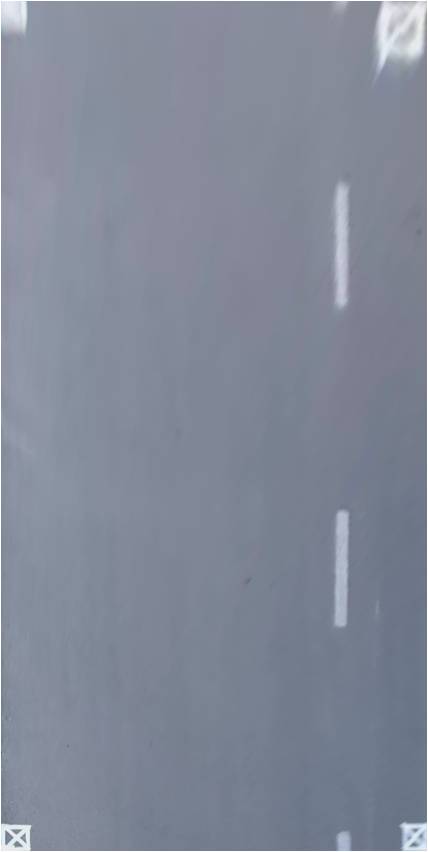} 
\end{center}
 \caption{Left: a view of the road, with zoom-in on the 4 calibration markers. The markers are
   identified and used to compute the calibration homography $\cH$. Right: after $\cH$ is applied to
   the part of the image surrounded by a square in the left panel, one gets a bird's eye view of
   this part of the road. }
 \label{fig:calibration}
\end{figure}

The calibration procedure is illustrated in Fig.~\ref{fig:calibration}. Note that when $\cH$ is
applied to the image, it transforms the image such that all the points on the reference plane are
brought to their correct position in space, while the mapping of other points depends on their
height relative to the reference plane.

\subsection{Image motion computation}
\label{sec:OF}

\begin{figure}[htb]
\begin{center}
\includegraphics[width=.85\linewidth]{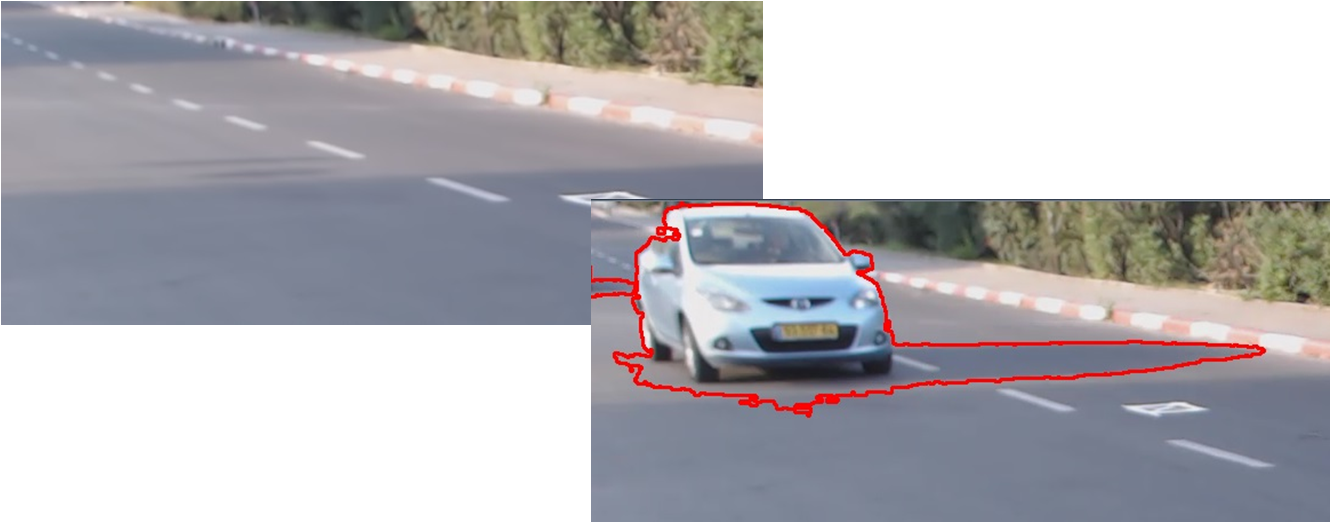} 
\end{center}
 \caption{Example of background subtraction: the frame on the right is compared to the background
   (frame on the left), and the difference is encircled by a red contour.}
 \label{fig:background}
\end{figure}

Here our task is to isolate the location of the license plate in each video frame, and track it
across numerous frames. This is done in a few steps:

\paragraph{Removal of empty frames:} 
we avoid heavy processing of empty video frames by using background subtraction
\cite{piccardi2004background}, as illustrated in Fig.~\ref{fig:background}. Frames that are judged
to resemble the background too much are removed from further processing.

\begin{figure}[bht]
\begin{center}
\begin{tabular}{lc} 
\vspace{0.2cm}
a) & \hspace{0.2cm} \includegraphics[width=.85\linewidth]{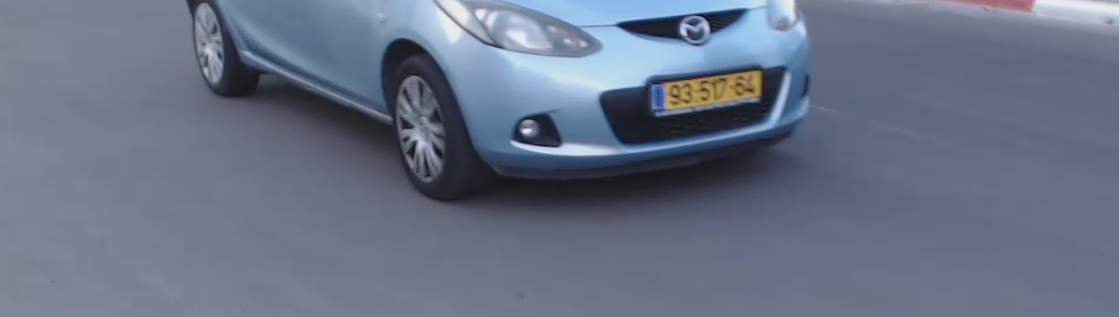} \\
\vspace{0.2cm}
b) & \hspace{0.2cm} \includegraphics[width=.85\linewidth]{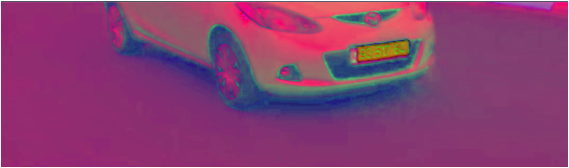} \\
\vspace{0.2cm}
c) & \hspace{0.2cm} \includegraphics[width=.85\linewidth]{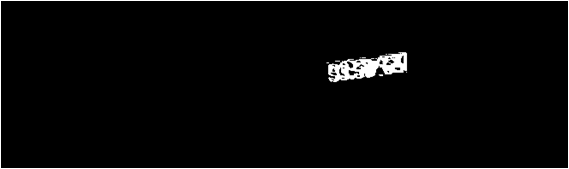} \\
\vspace{0.2cm}
d) & \hspace{0.2cm} \includegraphics[width=.85\linewidth]{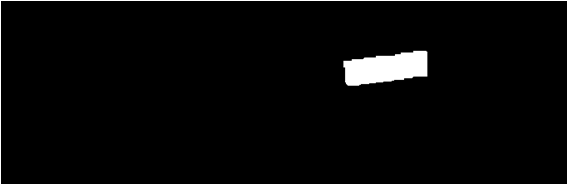} \\
\vspace{0.2cm}
e) & \hspace{0.2cm} \includegraphics[width=.85\linewidth]{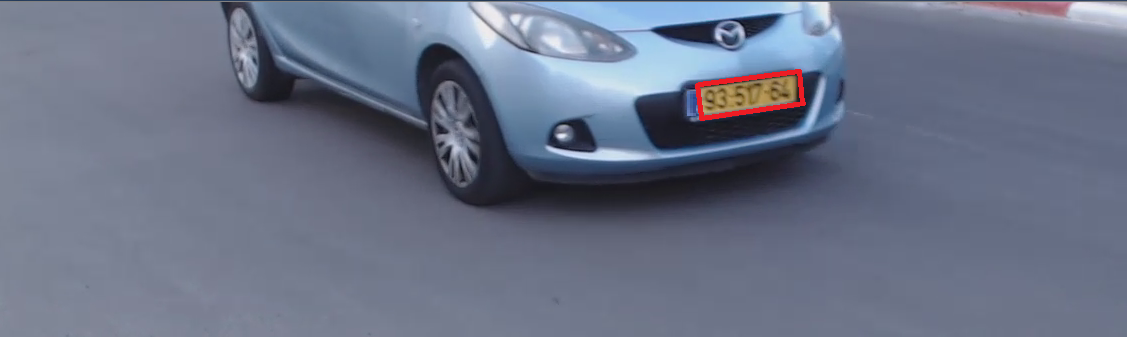} \\
\vspace{0.2cm}
f) &  \includegraphics[width=.5\linewidth]{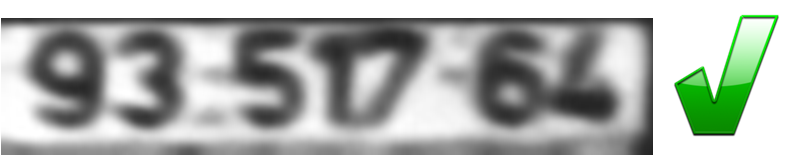} \\
\end{tabular}
\end{center}
 \caption{License plate segmentation. a) The original frame. b) The frame transformed to HSV color
   space. c) Pixels displaying the license plate color are isolated. d) Morphological operators are used
   to generate a clean segmented region. e) The position of the plate's bounding rectangle is sought
   around the boundary of the segmented area. f) The segmented area has been identified as a license
   plate by the SVM classifier.}
 \label{fig:segmentation}
\end{figure}

\paragraph{License plate segmentation:}
Segmentation follows a sequence of steps as illustrated in Fig.~\ref{fig:segmentation}. First, we
note that license plates are often characterized by some unique and easy to detect color, and
segmentation based on this unique color can be done rather reliably. We therefore transform the
images to HSV color space (Fig.~\ref{fig:segmentation}b), learn the appearance of a standard license
plate in this space, and use this to detect areas of the same color in all active frames. The image
is then binarized to remove all pixels of different color
(Fig.~\ref{fig:segmentation}c). Subsequently, morphological operators - dilation and erosion - are
used to unite close connected components and reduce noise (Fig.~\ref{fig:segmentation}d). Finally,
the occluding rectangular contour of the license plate is obtained in the surrounding of the
boundary of the segmented shape (Fig.~\ref{fig:segmentation}e).

\begin{figure}[htb]
\begin{center}
\includegraphics[width=.45\linewidth]{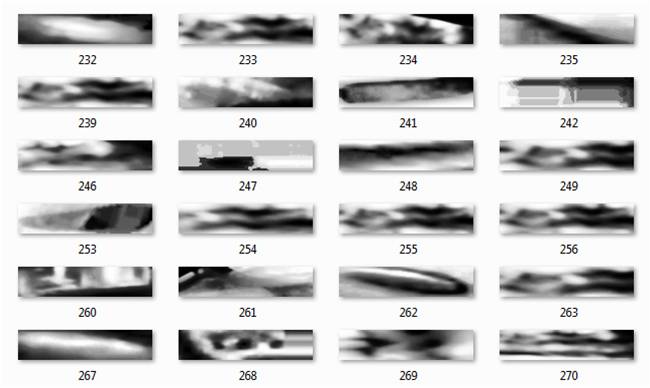} 
\hspace{.05\linewidth}
\includegraphics[width=.45\linewidth]{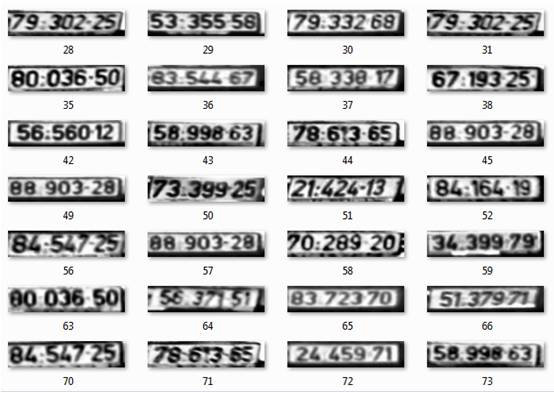} 
\end{center}
 \caption{Left: negative examples used to train an SVM classifier which decides whether a cropped
   region contains a license plate or not. Right: positive examples. }
 \label{fig:SVM}
\end{figure}

\paragraph{License plate classification:}
In order to decide whether the area segmented in the previous step is indeed a license plate, we
trained an SVM classifier using positive and negative samples obtained from video clips we had
collected as described in \cite{baggio2012mastering}. The samples were obtained using the following
procedure:
\begin{itemize}
\item
Rotate the image in order to compensate for the rotation angle of the region's occluding rectangle.
\item
Crop the rectangular region and resize it to $33 \times 144$ greylevel pixels.
\item
Apply histogram equalization.
\item
Vectorize the region and use it as a sample for the SVM classifier.
\end{itemize}
Some of the training examples are shown in Fig.~\ref{fig:SVM}. We divided the sample of 850 examples
into two parts, one for training and one for testing. Classification results, in trying to
distinguish regions which contain a license plate from other regions, had $0.45\%$ miss
rate. Example for positive identification is shown in Fig.~\ref{fig:segmentation}f.

\subsection{Reading the license number}
\label{sec:NN}

\begin{figure}[htb]
\begin{center}
\includegraphics[width=.52\linewidth]{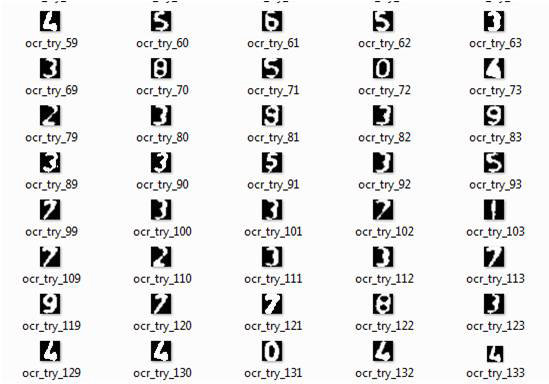} 
\hspace{.05\linewidth}
\includegraphics[width=.35\linewidth]{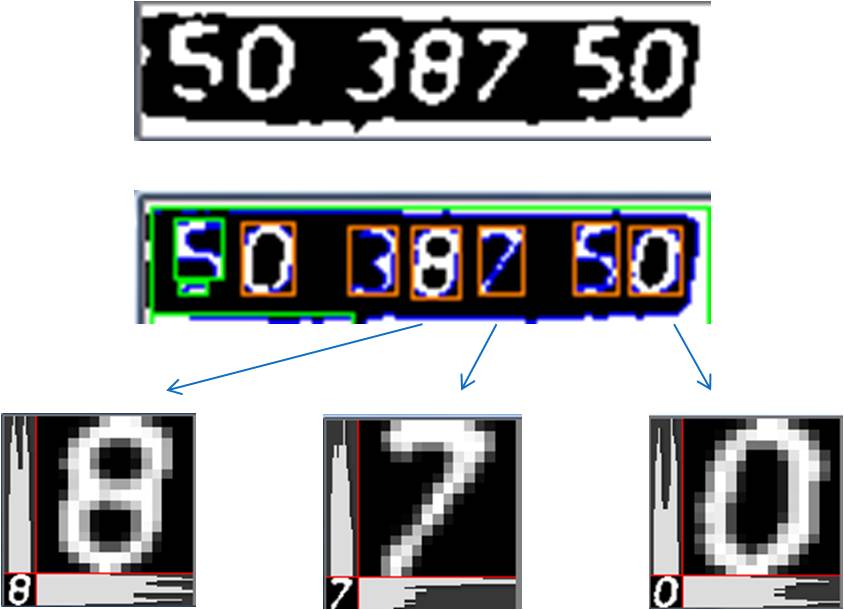} 
\end{center}
 \caption{Left: cropped characters used for training the ANN classifier. Right: illustration of the
   outcome of classification with the trained ANN. }
 \label{fig:OCR}
\end{figure}

Our task here is to read the license number from the image segment obtained in the previous step of
the algorithm, which has segmented the license plate from the rest of the image. OCR (optical
character recognition) has been the subject of much research in the last 40 years or so, where
Artificial Neural Networks (ANN) have emerged as one of the most effective methods for this task
(\eg \cite{parisi1998car}). For our purpose we trained an ANN with 3 layers and a Sigmoidal
activation function as described in \cite{baggio2012mastering} chapter 5. For training we used
characters that have been cropped from license plates we recognized in the videos we have collected
(see Fig.~\ref{fig:OCR}-left).  The trained ANN classifier first segments the image of the license
plate into individual characters, and then recognizes each character based on its pre-training (see
Fig.~\ref{fig:OCR}-right).

\subsection{Speed detection}
\label{sec:speed}

We start by tracking an interest point on the license plate for as long as possible. To this end we
use the corner detection algorithm described in \cite{shi1994good} to accurately find the corner of
the plate as illustrated in Fig.~\ref{fig:corners}.

\comment{
\begin{figure}[htb]
\floatbox[{\capbeside\thisfloatsetup{capbesideposition={left,bottom},capbesidewidth=.27\linewidth}}]{figure}[\FBwidth]
{\hspace{0.1cm}\includegraphics[width=1.1\linewidth]{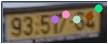}}
{\caption{Tracking the intersection of the wheel with the road can be difficult to do reliably.}
\label{fig:corners}}
\end{figure}
\end{figure}
}
\begin{figure}[htb]
\includegraphics[width=0.5\linewidth]{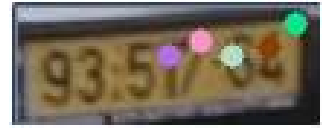}
\caption{Illustration of candidates for the upper right corner of the license plate; the green point
  is eventually selected as the position of the corner.}
\label{fig:corners}
\end{figure}

Speed detection proceeds as follows: Denote by $\bp_1$ and $\bp_2$ the image locations of a pair of
points in the tracking sequence. Project these points onto the road plane using homography $\cH$, to
obtain points $\cH \bp_1$ and $\cH \bp_2$ on the actual road plane.  Scale these projective
coordinates to obtain the corresponding Euclidean coordinates, and compute the distance between the
locations in order to estimate the {\em projected} Euclidean distance traveled by the interest
point. Using the known time that had passed between the frames in the tracking sequence, divide the
projected travel distance by the travel time to obtain an estimate for the speed of the license
plate. This speed is not the true speed of the license plate, because the interest point does not
move on the road plane but rather on a plane parallel to the road plane.

\subsection{Speed correction factor}
\label{sec:speed_correction}

Recall that $\cH$ defines the transformation (or homography) in $2D$ projective space between the
image plane and a real plane in the world ({\em the reference plane} on which the road lies. When
$\cH$ is applied to the image, it transforms all the points which actually lie on the reference
plane to their true location on this plane. But what happens to other points which do not lie on the
reference plane?

\begin{figure}[htb]
\begin{center}
\includegraphics[width=.85\linewidth]{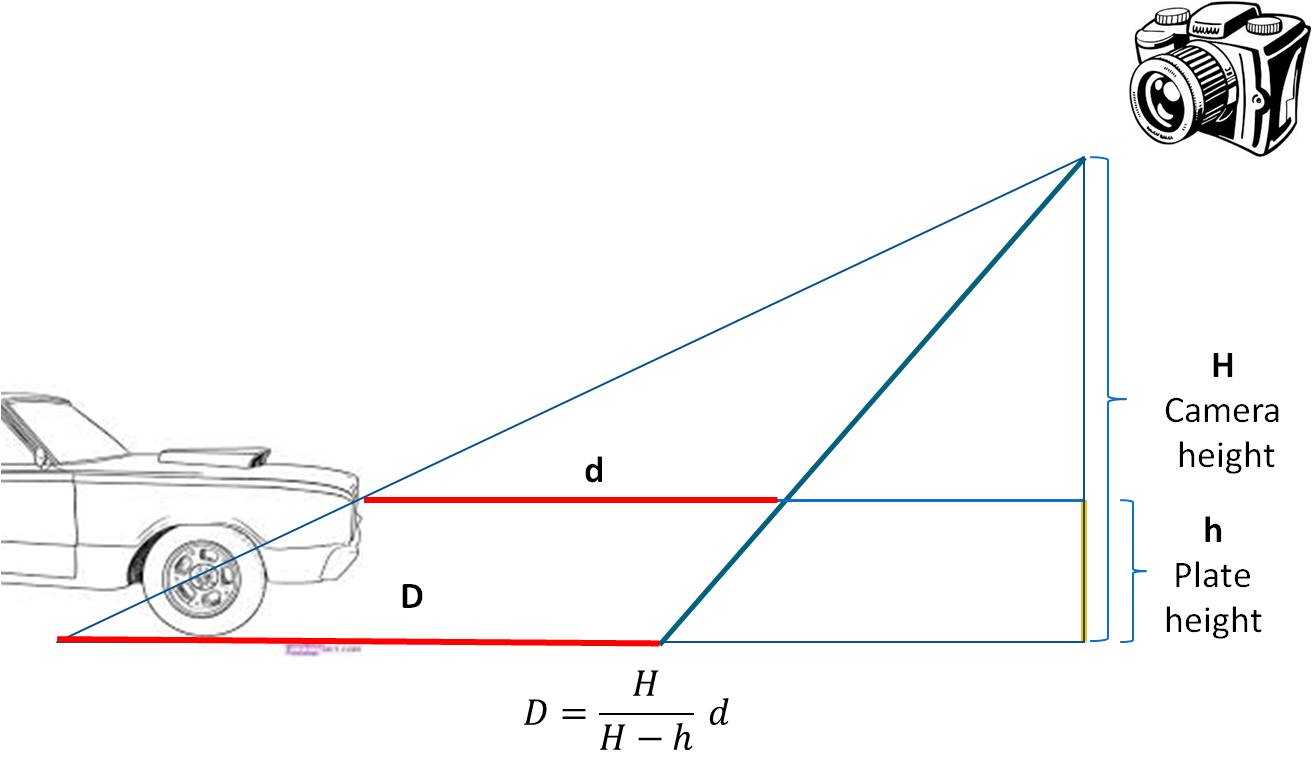} 
\end{center}
 \caption{The geometry of image formation after transforming the image by the homography which maps
   the image plane to the reference plane (which is the road plane). $\sD$ denotes the distance
   traveled by some point on the license plate after being projected to the reference plane. $\sd$
   denotes the real distance traveled by this point. $\sH$ denotes the height of the camera's center
   of projection above the road. $\sh$ denotes the height of the tracked point on the license plate
   above the road. }
 \label{fig:projection}
\end{figure}

After applying $\cH$ to the image, the combined image formation process can be imagined to be as
follows: the center of the pinhole camera remains as it has been, but the world in now projected
through this center onto a different plane, {\em the reference plane}, which is identical to the
image plane. This geometry is illustrated in Fig.~\ref{fig:projection}, which demonstrates what the
homography $\cH$ does to $3D$ points that do not lie on the reference plane: each such point is
effectively projected to the reference plane via the camera's projection center, as if the reference
plane is itself the imaging surface of the camera.

Recall that in the previous step of the algorithm, we computed the projected distance traveled by
some interest point on the license plate. This quantity is denoted in Fig.~\ref{fig:projection} by
$\sD$. The real distance traveled by the interest point is $\sd$. Therefore the correction
coefficient $\rho$, which brings the projected traveled distance of an interest point to the actual
distance traveled by this point, is
\begin{eqnarray}
\rho = \frac{\sd}{\sD} = \frac{\sH-\sh}{\sH}
\end{eqnarray}
where $\sh$ now denotes the height of the interest point over the road.

If using interest points which lie on the corners of the license plate, we can stop here when the
height of the license plate of the tracked car is known. This is the case when the  system has
access to a database of all licensed cars, which includes each car's license plate number and car
model.

\begin{figure*}[tb]
\begin{center}
\includegraphics[width=.3\linewidth, height=.17\linewidth]{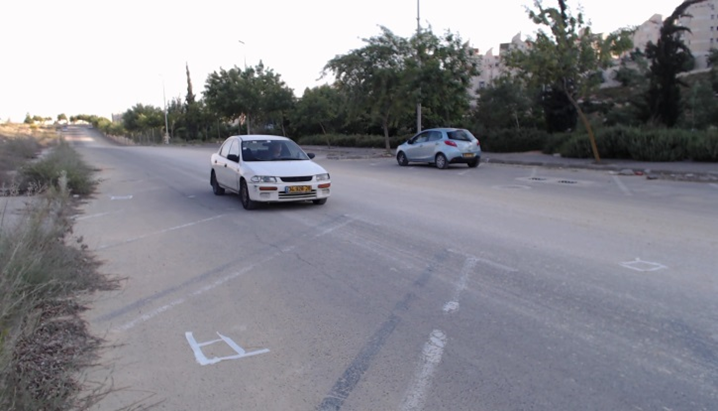} 
\hspace{.04\linewidth}
\includegraphics[width=.3\linewidth, height=.17\linewidth]{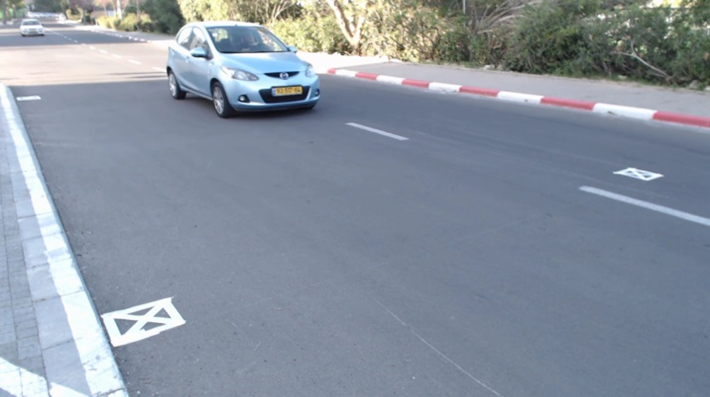} 
\hspace{.04\linewidth}
\includegraphics[width=.3\linewidth, height=.17\linewidth]{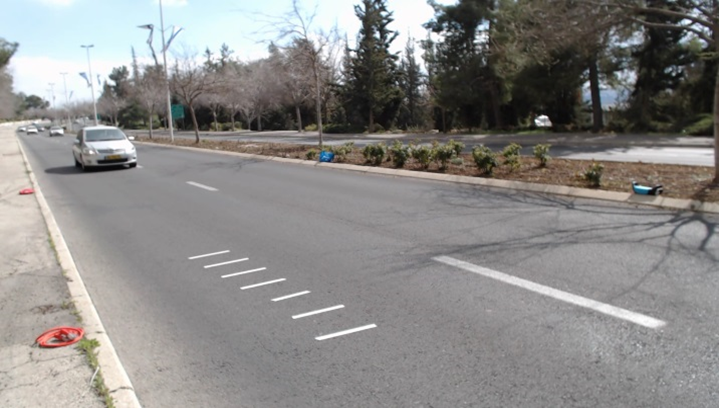} 
\end{center}
 \caption{Snapshots from several scenes used in our experiments.}
 \label{fig:scenes}
\end{figure*}

\begin{figure}[htb]
\begin{center}
\includegraphics[width=.85\linewidth]{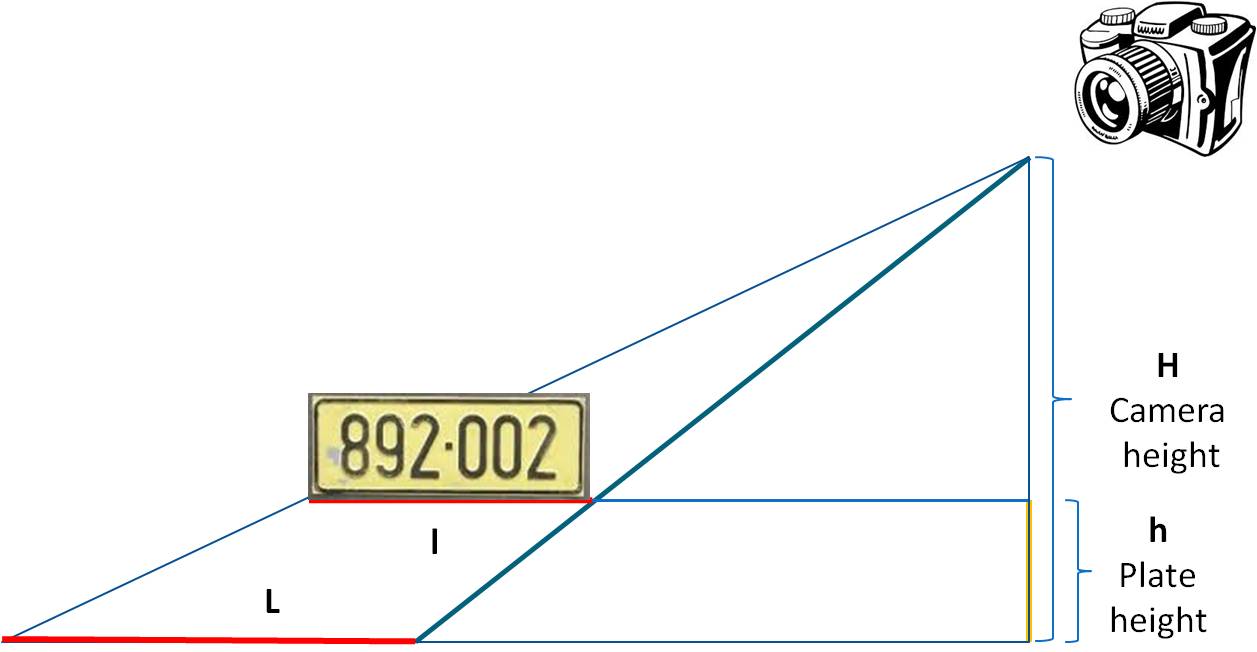} 
\end{center}
 \caption{Same geometry as Fig.~\protect\ref{fig:projection}, where $\slL$ denotes the projected
   length of the license plate on the reference plane, and $\sll$ denotes its fixed standard length. }
 \label{fig:plate-distortion}
\end{figure}

If this is not the case, we can take advantage of the fact that license plates follow a standardized
size and have a fixed length. Consequently the real length of any horizontal edge on the license
plate is known. In addition, these edges are parallel to the road plane, and therefore the
distortion in their size when measured on the reference plane is identical to the distortion of the
travelled distance, see Fig.~\ref{fig:plate-distortion}.

It therefore follows that\footnote{Errors may arise when the license plate is not mounted
  horizontally, and due to the difference between the projection of the top and bottom edges of the
  license plate. We tested robustness to these sources of error in our experiments.}.
\begin{eqnarray}
\rho = \frac{\sll}{\slL} = \frac{\rm\scriptstyle standard~license~plate~length}{\rm\scriptstyle measured~license~plate~length} 
\end{eqnarray}
where by ``measured license plate length'' we refer to the length of a horizontal edge on the
license plate which passes through the interest point.

\section{Experiments}
\label{sec:exps}

The components of the system we have built include a consumer camera and a laptop, see
Fig.~\ref{fig:system}. Specifically, the camera is Logitech Webcam C920. The laptop is Acer Timeline
U M5-481TG (Intel Core i5 3317U Processor 1.7GHz (3MB Cache) and 4 GB SDRAM RAM). The system was
tested with several cars driven in various speeds on different city roads. For illustration,
Fig.~\ref{fig:scenes} shows pictures from 3 of the scenes. In the supplementary material we include
movies showing license plates being detected and tracked in the corresponding scenes. Ground-truth
speed was measured by a GPS speedometer, due to the fact that most cars' speedometers are
intentionally biased by manufactures.

\begin{figure}[t]
\begin{center}
\includegraphics[width=.85\linewidth]{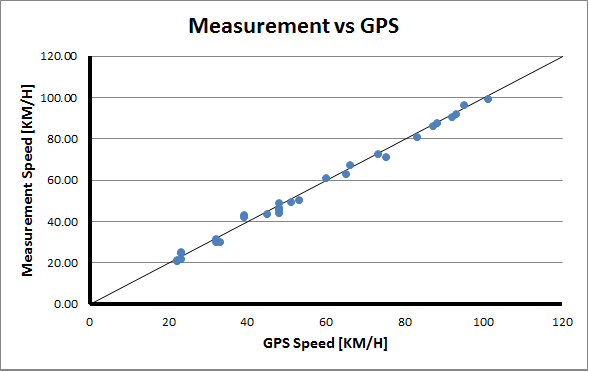} 
\bigskip
\includegraphics[width=.85\linewidth]{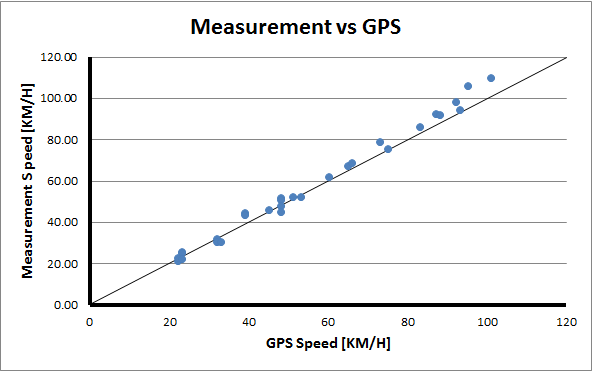} 
\end{center}
 \caption{The measured ground speed is plotted as a function of the actual GPS-measured ground
   speed, using method 1 (top) and method 2 (bottom). The solid line indicates the correct answer.}
 \label{fig:results}
\end{figure}

\begin{figure}[t]
\begin{center}
\includegraphics[width=.85\linewidth]{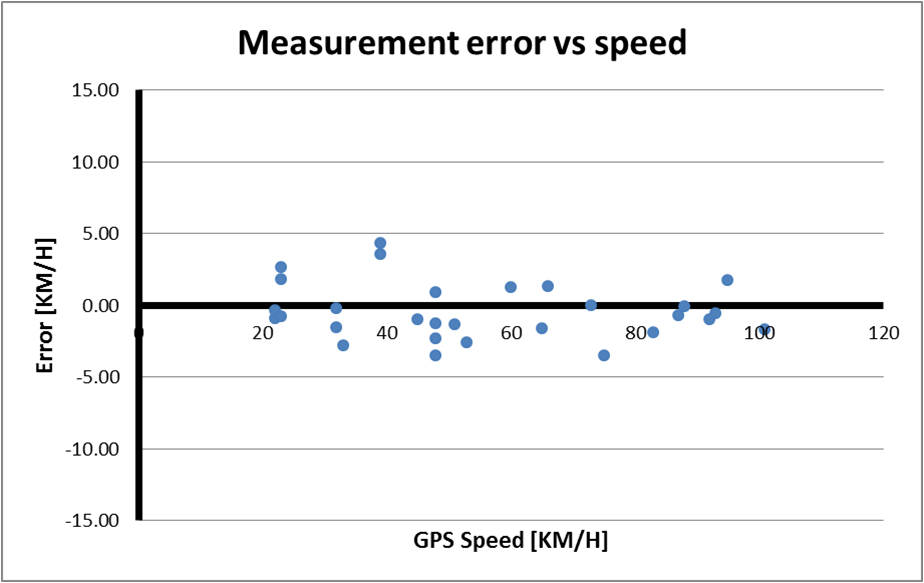} 
\medskip
\includegraphics[width=.85\linewidth]{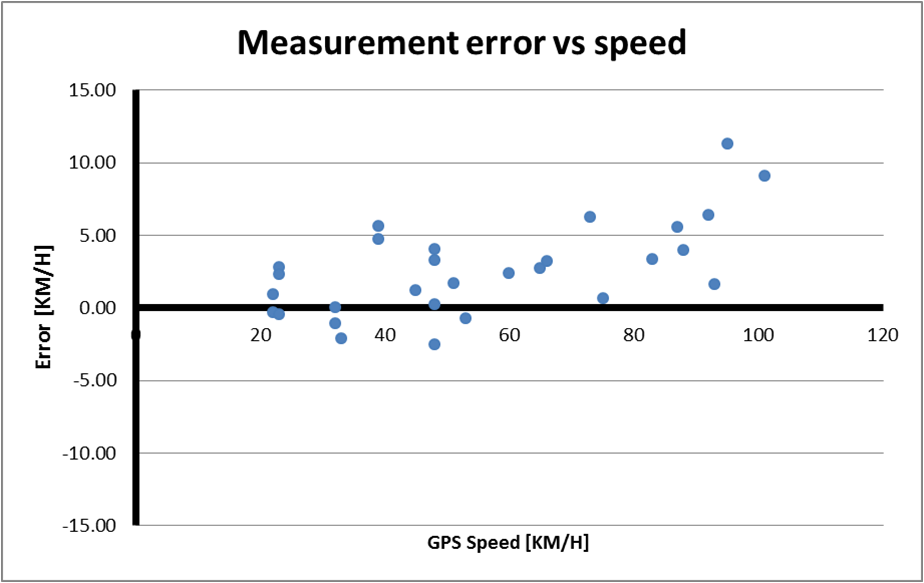} 
\end{center}
 \caption{Like Fig.~\protect\ref{fig:results}, showing the difference between the estimated speed
   and the true speed as a function of the true speed, using method 1 (top) and method 2 (bottom).}
 \label{fig:results-diff}
\end{figure}

The algorithm was implemented as described in Section~\ref{sec:method}; for each vehicle we measured
the correction factor $\rho$ from many frames, and took the median of the results.

As explained in Section~\ref{sec:speed_correction}, we used two method to compute the final speed of
the car's license plate. {\em Method 1} computes the correction factor $\rho$ while assuming that
the car's model, and consequently the height of the license plate, is known. {\em Method 2} computes
the correction factor $\rho$ directly from the foreshortening of the license plate caused by its
projection onto the reference plane. Figs.~\ref{fig:results}-\ref{fig:results-diff} show the
collected results of the ground speed estimation in all the different conditions based on the
different methods.

\section{Summary and discussion}

We described a system that computes the speed of moving vehicles from videos taken by a consumer
camera reliably and effectively. Our method is based on two assumptions: First, the existence of
some calibration markings on the road that are visible to the camera on occasion. Second, cars are
assumed to carry standard license plates mounted on the front of the car. In our future work we will
try to relax some of these assumptions, deal with a larger variety of license plates, improve the
OCR performance and achieve real-time performance.

{\small
\bibliographystyle{ieee}
\bibliography{paper}
}

\end{document}